\documentclass[letterpaper, 10 pt, conference]{ieeeconf}  % Comment this line out if you need a4paper

\IEEEoverridecommandlockouts                              % This command is only needed if 
\usepackage{booktabs}
\usepackage{multirow}
\usepackage{graphicx} 
\usepackage{amsmath}
\usepackage{amssymb}
\usepackage{xcolor}
\usepackage{hyperref}
\usepackage{booktabs}
\usepackage{tabularx}
\usepackage{pifont}
\usepackage{comment}
\usepackage{censor}
\usepackage{siunitx}
\usepackage{booktabs}
\usepackage{makecell}
\usepackage[normalem]{ulem}
\newcommand{\ofr}{ObjectFolderReal}
\newcommand{\ofb}{ObjectFolder Benchmark}
\newcommand{\ycbs}{YCB-Slide}

\newcommand{\sparsh}{Sparsh}
\newcommand{\digit}{DIGIT}
\newcommand{\gelsight}{GelSight}
\newcommand{\frthree}{FR3}
\newcommand{\flowtouch}{FlowTouch}
\newcommand{\flowmatching}{Flow Matching}
\newcommand{\ofrgel}{GelSight-Web} % ofr gelsight
\newcommand{\simgel}{GelSight-Sim} % sim gelsight
\newcommand{\simdigit}{DIGIT-Sim} % sim digit
\newcommand{\graspgel}{GelSight-Grasp} % grasp stability gelsight
\newcommand{\ycbdigit}{DIGIT-Web} % ycb digit
\newcommand{\selfcollected}{DIGIT-Ours} % self collected digit

\newcommand{\naiveft}{Naive Finetuning}
\newcommand{\domaincond}{Domain Conditioning}
\newcommand{\jitter}{Color Augmentation}
\newcommand{\optreset}{Optimizer Reset}
\newcommand{\sparshloss}{Tactile Loss}
\newcommand{\combined}{Combined}
\newcommand{\base}{Base}
\newcommand{\bgenc}{BG-Enc}
\newcommand{\bgsceneenc}{BG-Scene-Enc}
\newcommand{\bgstack}{BG-Stack}

\title{\LARGE \bf
\flowtouch: View-Invariant Visuo-Tactile Prediction
}
\StopCensoring % THIS
\author{\censor{Seongjin Bien$^{*1}$, Carlo Kneissl$^{*2,4}$, Tobias Jülg$^{*1}$, Frank Fundel$^{2,4}$, Thomas Ressler-Antal$^{2,4}$, \\Florian Walter$^{1, 3}$, Björn Ommer$^{2,4}$, Gitta Kutyniok$^{2,4,5}$ and Wolfram Burgard$^{1}$}
\thanks{\censor{$^*$Equal contribution}}%
\thanks{\blackout{$^{1}$Department of Computer Science~\& Artificial Intelligence, University of Technology Nuremberg, Germany. Contact: \texttt{seongjin.bien@utn.de}}}%
\thanks{\blackout{$^{2}$Ludwig Maximilian University of Munich}}
\thanks{\blackout{$^{3}$TUM School of Computation, Information and Technology, Technical University of Munich, Germany}}
\thanks{\blackout{$^{4}$Munich Center for Machine Learning (MCML)}}
\thanks{{\blackout{$^{5}$University of Tromso, DLR-German Aerospace Center}}}
}
\begin{document}
\bstctlcite{IEEEexample:BSTcontrol}

\maketitle
\thispagestyle{empty}
\pagestyle{empty}

%%%%%%%%%%%%%%%%%%%%%%%%%%%%%%%%%%%%%%%%%%%%%%%%%%%%%%%%%%%%%%%%%%%%%%%%%%%%%%%%
\begin{abstract}
% OLD VERSION FROM GRC Humans can predict tactile sensation from visual stimuli. This remarkable ability helps us to guide the way we manipulate objects and interact with our environment. However, tactile prediction is not an isolated capability that arises solely from vision. It draws upon prior experiences and models formed about the environment. Developing a similar capability for robots should thus not be viewed as a purely vision-based task, but rather as a problem that requires multiple modalities to achieve similar results. In this work, we propose FlowTouch, which builds upon these premises for tactile state prediction. It leverages 3D information of the target object to extract more information than would be available in an image alone, in order to achieve more robust tactile prediction capabilities that can generalize to a wider range of objects.

% Leave some comments for Florian on experiment results so that he can write the abstract in full better 
Tactile sensation is essential for contact-rich manipulation tasks. It provides direct feedback on object geometry, surface properties, and interaction forces, enhancing perception and enabling fine-grained control. An inherent limitation of tactile sensors is that readings are available only when an object is touched. This precludes their use during planning and the initial execution phase of a task. Predicting tactile information from visual information can bridge this gap. A common approach is to learn a direct mapping from camera images to the output of vision-based tactile sensors. However, the resulting model will depend strongly on the specific setup and on how well the camera can capture the area where an object is touched. In this work, we introduce \flowtouch, a novel model for view-invariant visuo-tactile prediction. Our key idea is to use an object’s local 3D mesh to encode rich information for predicting tactile patterns while abstracting away from scene-dependent details. \flowtouch{} integrates scene reconstruction and \flowmatching-based models for image generation. Our results show that \flowtouch{} is able to bridge the sim-to-real gap and generalize to new sensor instances. We further show that the resulting tactile images can be used for downstream grasp stability prediction. Our code, datasets and videos are available at \censor{\href{https://flowtouch.github.io/}{https://flowtouch.github.io}}

\end{abstract}

%%%%%%%%%%%%%%%%%%%%%%%%%%%%%%%%%%%%%%%%%%%%%%%%%%%%%%%%%%%%%%%%%%%%%%%%%%%%%%%%
% \section{TODO}
% \begin{itemize}
%     \item Change the bulletpoints to proper sentences
%     \item Get the necessary model details like token dimensions and model structures
%     \item Make a nice model overview diagram
%     \item Get the necessary pictures, namely: input-3PP, input-Gaussian, target-ground truth, inference-output, and validation data comparison
%     \item Get opinion on what to do for introduction + amount of literature review necessary - I would say keep it minimal, to less than 10 since we don't have that much space
% \end{itemize}

\section{Introduction}
% What the intro should mention:
% 1. Some humans blabla text as the intro for tactile, which also stresses that humans need more than just vision for tactile prediction
% 2. Visuo-tactile sensors
% 3. Recent advancements & works doen in this direction

% 5. Statement about static tactile sensation prediction is mroe about understanding object shapes in 3D, and that it makes more sense to use models with better priors about extracting these information instead of a general-purpose 2D image encoder 
% 6. Brief summary of our approach and using 2D to 3D models to leverage pretrained weights better, and achieve visual scene independence 
% 7. Contributions
%   - 3D-based approach is viable for tactile data generation
%   - Simulation data can be mixed in much more effectively through this
%   - Validate our approach through extensive experimentation

% ChatGPT iteration 1 
% 1, 2

% {\bf \color{red} WB: I really do not like the {\textsc{TextSC acronyms} I would only use them for abbreviations that were typeset in this way in the original papers}}

The sense of touch helps us interact with the world in our everyday lives. Imbuing robots with similar capabilities has been a longstanding goal. Progress in this domain has accelerated in recent years thanks to low-cost visuo-tactile sensors such as \gelsight{}~\cite{gelsight} and \digit{}~\cite{digit}, and the rapid advancement of vision processing techniques that complement these sensors. Still, efficiently integrating information from these sensors to boost robot performance remains an active field of research.

% Humans are born with high-resolution, ultra-sensitive tactile sensors that cover our entire body, especially in our hands, which help us interact with the world in our day-to-day lives. Imbuing robots with similar tactile capabilities has been a longstanding topic in the field. In recent years, visuo-tactile sensors such as GelSight~\cite{gelsight} and DIGIT~\cite{digit} have gained popularity thanks to their relatively low cost and high information density. Still, integrating the information from these sensors efficiently to boost robot performance remains an active field of research. 

Tactile data has been shown to improve performance
on complex manipulation tasks~\cite{chen_visuo-tactile_2022, hansen_visuotactile-rl_2022, higuera_perceiving_2023,calandra_feeling_2025}, as they provide fine-grained information about the interaction that is not directly observable through vision alone. However, tactile sensors only provide relevant output during physical contact. As a result, at the beginning of a manipulation task, robots must rely primarily on vision to perceive and reason about their environment. To effectively integrate these complementary modalities, a robot should anticipate the expected tactile feedback before contact, reflecting the close alignment of vision and touch observed in the brain~\cite{Keysers.2010, Hedger.2025}. Such a prediction enables more informed action planning and smoother transitions from non-contact to contact interaction.

% Several works have explored this direction by leveraging recent advances in deep learning-based visual processing to train generative models that predict tactile sensor outputs conditioned on camera images~\cite{wu_vitacgen_2025,gao_objectfolder_2021,li_connecting_2019}. However,
% % ... data expensive... generalize poorly... 
% % these approaches suffer from the sheer magnitude of data that needs to be collected to generalize across different scene and object variances.
% the ability of humans to make cross-modal predictions is likely not based solely on visual processing. We know intuitively that the tactile sensation of grasping an apple does not change regardless of whether the object is located in a well-lit kitchen or buried under a pile of groceries in a bag. This suggests that we use visual stimuli as cues to draw on object priors from previous experiences and our internal model of the world, which is inherently object-centric and invariant to the visual context. Given this observation, we posit tactile prediction requires an object-centric representation---namely, geometric information.

% wolfram: motivate why we are going to meshes, focus on why -> learning efficiency, less data (direct vision to touch would be very expensive, a lot of data needed)
Several works have explored this direction by leveraging recent advances in deep learning-based visual processing to train generative models that predict tactile sensor outputs conditioned on camera images~\cite{wu_vitacgen_2025, gao_objectfolder_2021, li_connecting_2019}. However, learning a direct mapping from camera space to tactile space makes incorporating prior information difficult, requiring huge amounts of data across different scenes and objects to ensure good performance and generalization. Simulation can only partially mitigate this issue as both RGB renderings and tactile sensor models introduce a domain shift relative to real-world data. However, only a small fraction of visual features is relevant for tactile prediction, which mainly depends on geometric features. Including object-centric geometry information can therefore mitigate this issue by abstracting away from visual details not relevant for tactile sensing. However, this is difficult for purely vision-based end-to-end models.

In this work, we propose \textit{\flowtouch{}}, a vision-to-touch framework that addresses this gap by explicitly leveraging geometric information to condition a \flowmatching–based generative model for predicting static tactile sensations from camera images. By lifting target objects from 2D observations into 3D representations using scene mesh generation foundation models, \flowtouch{} achieves increased robustness to viewpoint variations and scene-level changes.

\begin{figure}[t]
  \vspace{5.1pt}
  \centering
    \includegraphics[width=0.9\linewidth]{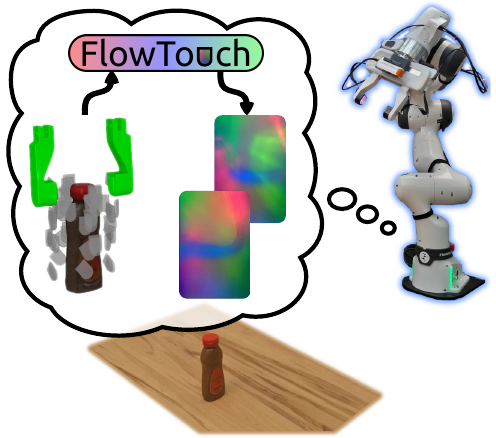}
        \caption{\flowtouch{} at a glance: The robot first looks at the object and creates a mesh with scene mesh generation foundation models. Given the particular touch point on the mesh and the static background image of the tactile sensor, \flowtouch{} then predicts the resulting tactile image.}
        \label{fig:eyecandy}
\vspace{-1.5em}
\end{figure}

\pagebreak
\textbf{In summary, we make the following contributions:}
\begin{enumerate}
    \item We introduce a generative framework conditioned on geometry that predicts tactile signals without requiring exploratory robot motion beyond initialization.
    \item We propose a simulation-to-real training method based on the mesh data of geometric primitives obtained via simulation and show that it improves output quality, substantially reducing the need for costly real-world data collection.
    \item We demonstrate our framework's capability by empirically verifying its strong generalization across scene variations and sensor instances, and its ability to generate relevant information for downstream manipulation tasks through grasp stability experiments.% 
\end{enumerate}

\section{Related Work}
\subsection{Vision-to-Tactile Image Prediction}
Previous work on robotic vision-to-tactile image prediction typically conditions a generative model on camera images captured during tactile interaction to produce the resulting visuo-tactile signal. Both VisGel~\cite{li_connecting_2019} and ViTacGen~\cite{wu_vitacgen_2025} follow this approach, encoding interaction images and directly using the resulting representation. Because such conditioning inherently entangles scene and viewpoint with tactile-specific information, these methods often struggle to transfer beyond a benchmark setup. In contrast, \mbox{\ofb{}}~\cite{gao_objectfolder_2023} uses a pre-contact image from the same viewpoint as the target visuo-tactile observation, reframing the prediction as an inpainting task where much of the required geometric information is already present. Both paradigms require the robot to move close to the contact pose to acquire suitable inputs, which limits their practical use. 

The closest parallel to our proposed work is Touching a NeRF~\cite{touching_nerf_2022}, which trains a neural radiance field (NeRF) per object to render RGB and depth images from perspectives similar to the \ofb{} to be used as conditioning for a generative model. 
However, this approach requires object-specific training data for each novel instance, limiting its scalability. We address these issues directly by leveraging foundation models to lift objects from image to geometric space, thereby improving robustness to novel objects and scene variations, and motion-free input acquisition for tactile prediction.

\subsection{Mesh-based Tactile Image Generation in Simulation}

Visuo-tactile sensors such as \gelsight{}~\cite{gelsight} and \digit{}~\cite{digit} capture the deformation of a silicon surface upon contact. Several sophisticated simulation methods have been introduced to enable the generation of tactile images at scale. TACTO~\cite{Wang2022TACTO} emulates contacts and renders tactile images using off-the-shelf rendering and simulation engines. Other approaches combine finite element methods with either rendering engine-based~\cite{tacflex_2025,tacchi_2023} or data-driven methods~\cite{difftactile_2024,si_taxim_2021} to simulate both the tactile image and the deformation of the soft silicon pad to achieve higher fidelity. 
However, simulation data often suffer from non-trivial sim-to-real gap, preventing their direct application in real-world scenarios without significant effort~\cite{Chen.2022, Caddeo.2024}.
Our model leverages simulation data generated for large-scale pretraining and uses learning-based adaptation on real data to narrow the sim-to-real gap, enabling more robust tactile image generation.

\subsection{Tactile Datasets}

Existing tactile datasets combine touch with visual observations and, in some cases, object geometry and pose supervision. ObjectFolder~1.0~\cite{gao_objectfolder_2021} and 2.0~\cite{gao_objectfolder_2022} provide a multimodal dataset comprising over 1000 objects in neural implicit representations, albeit with simulated tactile signals. The succeeding \ofr{}~\cite{gao_objectfolder_2023} provides paired camera, tactile, and robot pose data for 100 real-world objects, while \ycbs{}~\cite{suresh2022midastouch} includes real and simulated \digit{} data with accurate sensor poses and ground-truth meshes. In contrast, Touch-and-Go~\cite{yang_touch_2022} and Touch Vision Language~\cite{tvl_2024} focus on visual–tactile pairing without precise pose or geometric supervision.

For our proposed approach, explicit geometry and sensor pose are crucial for localizing touch on object surfaces. Although \ofr{} and \ycbs{} provide this information, they offer limited pose and contact diversity, with most surface points sampled only once. We, therefore, introduce a large-scale simulation data generation pipeline that systematically varies contact locations and relative sensor poses to complement existing datasets.

\begin{figure*}[t]
  \centering
  \vspace{5.1pt}
  \includegraphics[width=\textwidth]{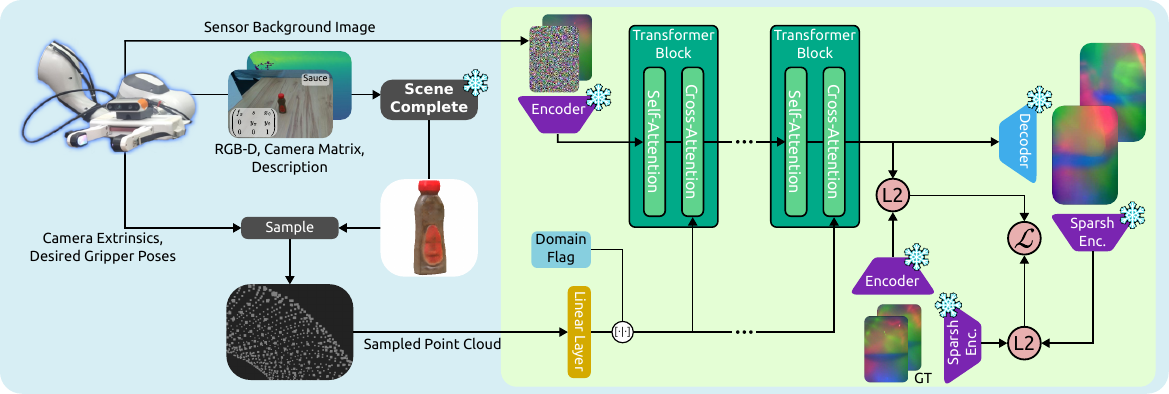}
  \caption{
    The \flowtouch{} architecture: The blue area shows the image-to-PCN sampling pipeline, while the green area shows the generative model's structure. Snowflakes indicate components with frozen weights.
  }
  \label{fig:arch}
  \vspace{-1em}
\end{figure*}

\section{Methodology}
% Things to talk about in methodology
% 1. Preliminaries -- a small write-up about how tactile sensors work
% 2. Engineering part: calibrated cam + robot, scene -> segment -> 3d and pose estimation, mesh cutout and VAE
% 3. Learning part: Outline the sim data generation pipeline and data mixing, and the training architecture
\subsection{Preliminaries}
% 1. Flow matching - conditional
% 2. Visuo-tactile sensors?

We address the problem of predicting a tactile sensor image $\mathbf{x} \in \mathbb{R}^{H \times W \times 3}$ based on a set of given contextual signals $\mathbf{c} = \{{\mathbf{m}, \mathbf{b}} \}$, where $\mathbf{m} \in \mathbb{R}^{N_\text{pcn} \times 6}$ is a point cloud of 3D positions and their respective normals (PCN) randomly sampled around the contact point on the mesh (with $N_\text{pcn}=5000$ points) and $\mathbf{b} \in \mathbb{R}^{H \times W \times 3}$ is a background image from the same tactile sensor captured in the base state without any contact. We learn the conditional distribution $p(\mathbf{x} \mid \mathbf{c})$ via a latent \flowmatching{} objective~\cite{lipman2023flowmatchinggenerativemodeling} on the compressed representation of $\mathbf{x}$.

\subsection{\flowtouch{}}
~\autoref{fig:arch} shows an overview of \flowtouch's architecture. The method consists of two components: the image-to-PCN sampling pipeline, and the \flowmatching{} model that generates the tactile image conditioned on the input PCN.
\vspace{0.3em}

\subsubsection{Sampling the PCN}\label{sec:meth_pointcloud}
In order to sample the PCN around the desired contact points on the target object, we first create a mesh of the object that is aligned with the robot coordinate frame to properly position the contacts from the robot gripper's desired poses. We employ SceneComplete~\cite{scenecomplete} to achieve this. The framework takes in RGB-D images, the camera intrinsic matrix, and a language description of the target object. It then performs object segmentation and image inpainting before feeding the results into a 2D-to-3D model, whose output is scaled and aligned to the original depth image using a 6-D pose estimation model. 

With the aligned mesh, we can query the desired grasp pose and calculate the resulting poses of the tactile sensors mounted at the robot gripper's fingertips. To ease the sampling process, we use MuJoCo~\cite{mujoco} as follows: as MuJoCo creates a convex hull around a given collision-enabled geometry, we first decompose the output mesh into several convex submeshes using CoACD~\cite{wei2022coacd}, and check for contacts only on those meshes, while keeping the original mesh for sampling. We then initialize a MuJoCo scene with the object mesh, its decomposed convex meshes, and the contact pad meshes of a pair of tactile sensors. To account for minor alignment errors, the contact pads are initialized at the gripper fingers' open positions and adjusted along the gripper's closing direction until both pads are penetrating any of the convex submeshes by \qty{3}{mm}. Finally, we sample the PCN from the intersection of the mesh surface with the visualization areas of the two pads, which are smaller than the pads themselves. This PCN is used to condition the transformer-based image-to-image generative model.
% \vspace{0.3em} % This helps to more easily distinguish the subsubsections 

\subsubsection{Latent Encoding}
Inspired by the success of Stable Diffusion \cite{rombach2022high-res}, all image quantities are first compressed to a latent space using a frozen, tiny autoencoder~\cite{bohan2023taesd}
$\mathcal{E}: \mathbb{R}^{H \times W \times 3} \to \mathbb{R}^{(H/8) \times (W/8) \times 4}$, reducing spatial resolution by a factor of 8 and projecting to a 4-channel latent representation. We denote the latent encoding of the target tactile image as $\mathbf{z} = \mathcal{E}(\mathbf{x}) \in \mathbb{R}^{h \times w \times 4}$ and the decoder as $\mathcal{D}$.
% \vspace{0.3em} % This helps to more easily distinguish the subsubsections 

\subsubsection{\flowmatching{} Objective}
We adopt the Conditional \flowmatching{}~(CFM) framework~\cite{lipman2023flowmatchinggenerativemodeling} with a linear interpolant and define the forward process as a straight path from noise to data:
\begin{equation}
   \mathbf{z}_t = (1 - t)\mathbf{z}_0 + t\cdot\mathbf{z}_1, %\qquad t \in [0, 1], 
\end{equation}
where  $t \in [0, 1]$, $\mathbf{z}_0 \sim \mathcal{N}(\mathbf{0}, \mathbf{I})$ is a standard Gaussian random variable and $\mathbf{z}_1$ is a sample from the latent target distribution. A U-Net style Vision Transformer \cite{Dosovitskiy_2020_AnImage} denoted by $v_\theta(\mathbf{z}_t, t, \mathbf{m})$ is trained to regress the conditional velocity $u_t(z_t \mid z_1) = z_1 - z_0$ with the CFM objective
% $$\mathcal{L}_{CFM}(\theta) = \mathbb{E}_{t, \mathbf{z}_0,\mathbf{z}_1} \left[ \left| v_\theta(\mathbf{z}_t, t, \mathbf{m}) - (\mathbf{z}_1 - \mathbf{z}_0) \right|^2 \right].$$
\begin{equation}
    \mathcal{L}_{CFM}(\theta) = \mathbb{E}_{t, \mathbf{z}_0,\mathbf{z}_1} \left[ \left| v_\theta(\mathbf{z}_t, t, \mathbf{m}) - (\mathbf{z}_1 - \mathbf{z}_0) \right|^2 \right].
\end{equation}
At inference time, the ordinary differential equation \mbox{$\dot{\mathbf{z}}(t) = v_\theta(\mathbf{z}_t, t, \mathbf{m})$} is integrated from $t=0$ to $t=1$ using the Euler method with $K$ uniform steps.
% \vspace{0.3em} % This helps to more easily distinguish the subsubsections 

\subsubsection{Background Image as Channel-Stacked Context}
\label{subsub:back_img_as_ch_st_ctxt}
The background image $\mathbf{b}$ encodes the sensor's baseline appearance in the absence of contact and serves as a spatial prior. Its latent $\mathbf{b}_z = \mathcal{E}(\mathbf{b}) \in \mathbb{R}^{h \times w \times 4}$ is encoded once per sample and concatenated channel-wise with the noisy tactile latent:
% $$\tilde{\mathbf{z}}_t = [\mathbf{z}_t | \mathbf{b}_z] \in \mathbb{R}^{h \times w \times 8},$$
\begin{equation}
    \tilde{\mathbf{z}}_t = [\mathbf{z}_t | \mathbf{b}_z] \in \mathbb{R}^{h \times w \times 8}
\end{equation}
We use $[\cdot |\cdot]$ to denote concatenation along the channel dimension. This formulation gives the backbone direct pixelwise access to the background at every integration step without additional cross-attention overhead, and preserves exact spatial registration between the deformation field and the sensor baseline.
% \vspace{0.3em} % This helps to more easily distinguish the subsubsections 

\subsubsection{Cross-attention Mesh Conditioning}
The sampled PCN of shape $\mathbf{m} \in \mathbb{R}^{N_\text{pcn} \times 6}$ are linearly projected by $\mathbf{W_\text{mesh}} \in \mathbb{R}^{6 \times {d_k}}$ to form the cross-attention context $\mathbf{mW_\text{mesh}}$ where $d_k=768$ is the hidden dimension. In each transformer layer, the spatial tokens $\mathbf{X}$ serve as queries while the mesh context provides keys and values, allowing every spatial location to attend over all $N_\text{pcn}$ points. Thus, our attention block looks as follows:
\begin{equation}
    \text{CrossAttn}(\text{SelfAttn}(\mathbf{X}), \mathbf{mW_\text{mesh}}).
\end{equation}

\subsubsection{Domain Adaptation}\label{sec:domain_adaptation}
% - mixing of data sources and avoid cathastripic forgetting
% - domain condition flag
In order to bridge the domain gap between simulated sensor values from our synthetic Taxim-generated dataset and datasets recorded with physical sensors, we employ a two-stage learning approach:
First, we pretrain only on synthetic data for 30k steps with batch size $128$.
It is about 100 times larger than our real-world dataset and consists of different basic geometric shapes and our two sensor types, as detailed in \autoref{sec:sim_data}.
Afterwards, we start a finetuning phase for another 20k steps where we mix in real data but only weigh it with 20\%. The remaining 80\% is still filled with synthetic data to avoid catastrophic forgetting.
We refer to this strategy as \textit{\naiveft} and use it as our baseline to ablate further domain adaptation methods on top.

% domain conditioning
To help the model generalize to core concepts shared across domains, we also implemented a domain conditioning token, a standard technique for domain adaptation~\cite{daume2007frustratingly,johnson2017google}.
We add a simple domain flag $dom\in\{0, 1\}$ (0~for synthetic and 1 for real) which is used to retrieve a learnable embedding $\mathbf{W}_\text{dom}\in\mathbb{R}^{d_k}$
% which is added onto the time embedding.
% rojected into token space by $\mathbf{W}_{dom}\in\mathbb{R}^{2\times d_k}$
and appended to the PCN information used in cross attention:
% Our attention block then looks as follows:
\begin{equation}
    \text{CrossAttn}(\text{SelfAttn}(\mathbf{X}), [\mathbf{mW_\text{pcn}} | \mathbf{W}_\text{dom}])
\end{equation}
This training technique is referred to as  \domaincond{}.

% optimizer reset
We also experiment with resetting the optimizer state and dropping its learning rate by a specified factor.
This avoids updating the weights in the wrong direction with the AdamW momentum of the pretraining dataset when finetuning starts.
We refer to this as \optreset{}.

% \subsubsection{\jitter{}}
% Overfitting can be a big problem in domain adaption especially when one domain has orders of magnitude less data. The model might have a tendency to just remember the training data if no regularization methods are applied. To break this behavior data augmentations such as adjusting the image's hue, saturation and brightness artificially increase the dataset size and make it, thus, easier for the model to concentrate on concepts and generalize. We also experiment with augmentations techniques during training and refer to such as \textit{\jitter{}}.

Overfitting is a major challenge in domain adaptation when one domain has significantly less data, causing models to merely memorize the training set. To encourage generalization, we experiment with color augmentations (hue, saturation, brightness)---referred to as \textit{\jitter{}}---to artificially expand the dataset and mitigate this behavior.

% Traditionally, data augmentations such as adjusting the hue, saturation or brightness of the image help the model to generalize and prevent overfitting and memorizing the training dataset to a specific domain that is overrepresented during training.
% which was dominantly seen in training.
% Therefore, we include it as a training method to avoid domain bias.
% We refer to this training technique as \textit{\jitter{}}.
% we though it would help with different sensors and sensor instances (mention all the parameters)

% \subsubsection{\sparshloss{}}
The output of visuo-tactile sensors depends largely on the specific hardware used, and thus, two identical touches can yield completely different visual readings across different hardware.
To mitigate this problem, we incorporate \sparsh{}~\cite{higuera2024sparsh}, a self-supervised encoder for tactile predictions. It is able to produce compressed embeddings that retain most of the relevant information of the original tactile image, such as geometry, normal and shear forces, across a wide variety of sensor types.
We hypothesize that a good tactile prediction from our model should be close to the ground truth in the \sparsh{} embedding space, as the embeddings retain information relevant to downstream tasks while ignoring less relevant data, such as sensor-specific noise.
Thus, we include a loss objective to enforce that our predictions are consistent in the \sparsh{} embedding space.

More formally, let $\mathcal{E}_\text{\sparsh} : \mathbb{R}^{H \times W \times 3} \to \mathbb{R}^{h' \times w' \times 6}$ denote the static \sparsh{} encoder.
%which takes as input two tactile images with temporal stride.
%As we focus entirely on a static environment, we simply feed in the same image twice, thus encoding no dynamical change.
In order to include this perception-based loss in our \flowmatching{} objective, we generate images with a single prediction from our \flowmatching{} model.
For this, we reuse the $\mathbf{z}_t$ computed for the \flowmatching{} loss and compute $\hat{\mathbf{z}}_1 = \mathbf{z}_t + (1-t) \cdot v_\theta(\mathbf{z}_t, t, \mathbf{m})$.
The Sparsh-based perceptual loss objective can then be formulated as
\begin{equation}
\mathcal{L}_\text{\sparsh} = 
\Vert \mathcal{E}_\text{\sparsh}(\mathcal{D}(\hat{\mathbf{z}}_1)) -\mathcal{E}_\text{\sparsh}(\mathbf{x}_1) \Vert_2,
\end{equation}
where $x_1$ denotes the ground truth tactile image.
\begin{equation}
\mathcal{L}_{FT} = \mathcal{L}_{\text{CFM}} + \mathcal{L}_{\text{\sparsh}}.
\end{equation}
We refer to this training technique as \textit{\sparshloss{}}.

% where $\beta$ and $\gamma$ are hyperparameters to weight the losses.
% that are adjusted based on the \textbf{source data type? training phase?}, discussed in \autoref{sec:exp}. 
% \vspace{0.3em}

% \subsubsection{Conditional flow matching}
% For the Image to image task we are using conditional flow matching
%     \begin{itemize}
%         \item It has achieved state-of-the art performance in generative modeling
%         \item with a comparatively low latency
%         \item Improved, training stability compared to GANs for example in particular they do not suffer from mode collapse 
%     \end{itemize}

% \subsubsection{Transformers and cross attention}
% \begin{itemize}
%     \item To model the trainable velocity field we are resorting to transformers, specifically ViT
%     \item Cross attention
% \end{itemize}

% MappingNetwork: input -> RMSNorm -> ..., but we don't use MappingNetwork I think
% Input to the backbone (i.e. centran Transformer blocK) uses Pos2D as the position arg, which is used as input to pos_encoding
% pos -> theta -> apply_emb(q, theta)
% pos_cross -> theta_cross -> apply_emb(k, theta)
% Backbone itself is 12 blocks of GenericCrossTransformerLayer
    % repeatedly self-attention on input x, then cross-attention on conditioning
    % pos_encoding = Rope2D
    % projection = Patch2D
% Image encoder-decoder is TinyAutoencoderKL, with weights madebyollin/taesd
% Frank said: RoPE is basically a standard at this point, they don't show it in their diagrams
%Point cloud into TripoSGVAE/Niessner

\subsection{Synthetic Data Generation}\label{sec:sim_data}
% Not sure about the ordering. Model -> Sim -> Calculating? 
\begin{figure}[t]
  \vspace{5.1pt}
  \centering
    \includegraphics[width=0.8\linewidth]{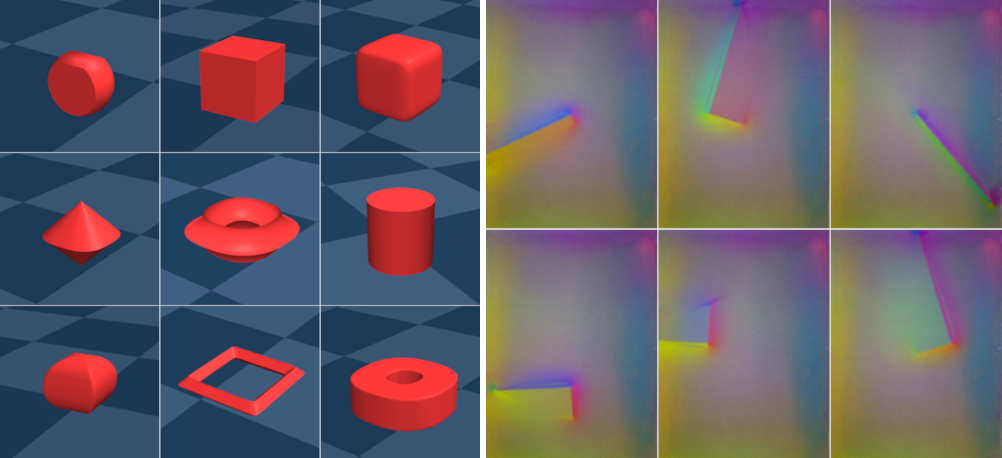}
        \caption{\textit{Left:} A selection of primitive geometries used for generating the simulation data. \textit{Right:} Taxim \gelsight{} images rendered from a single contact point (corner of a cube), with local translations and rotations.}
        \label{fig:mj_geoms}
\end{figure}
% The text alone is about a column's worth, which should be enough
% Need to mention somewhere that we are not focusing on texture, but larger geometric shapes

% More concise version
A major benefit of working directly in geometric space is the considerably reduced sim-to-real gap compared to image-based approaches. To fully leverage this advantage, we design a simulation data generation pipeline that captures a wide range of local surface geometries that visuo-tactile sensors may encounter in the wild.

We integrate Taxim~\cite{si_taxim_2021} into \mbox{MuJoCo}~\cite{mujoco}, leveraging its parametrizable primitive geometries and built-in contact handling to generate diverse shapes efficiently. By varying geometric parameters, we can construct a large library of objects (\autoref{fig:mj_geoms}, left) whose properties reflect those observed in existing tactile datasets and household items. Rather than sampling surfaces uniformly---which would over-represent flat regions---we explicitly select geometrically informative locations (e.g., edges and corners) to maintain a balanced coverage. At each location, we apply combinations of small local translations and rotations on the sensor to simulate realistic contact variations, and render the tactile image and sample the contact PCN (\autoref{fig:mj_geoms}, right). This process produces over 100k unique samples in under an hour on a consumer-grade laptop (i9-13900HX), but could be substantially sped up through a GPU implementation.

Because PCNs sampled from such synthetic geometries are perfectly smooth, we inject multi-scale Gaussian noise along sampled normals to mimic reconstruction artifacts typical of learned mesh generation. Additionally, to reduce sensor-specific bias, we randomize the background image for each render.
\section{Results}\label{sec:results}

% Primitives: box, sphere, supertorus, wedge, double ring
% Training data
% gelsight sim -> 262151, gelsight real -> 887
% digit real (YCB) -> 2578, digit sim -> 177319

We evaluate FlowTouch in three experiments. In \autoref{sec:result_design}, we validate the architectural design choices from the previous section and their effect on generated image quality. In \autoref{sec:result_training}, we assess the impact of data augmentations and training techniques. Finally, in \autoref{sec:result_grasp}, we test whether FlowTouch-predicted images support downstream grasp stability prediction. We use several datasets, summarized in \autoref{tab:datasets}. \ofrgel{} and \ycbdigit{} denote the ObjectFolderReal and YCB-Slide datasets downloaded from the web, respectively, which provide ground truth meshes and sample positions for extracting the PCNs required by our model. We preprocess the datasets to align tactile images with mesh poses, filter out large deviations, and create object-level train/validation splits. We also generate large simulation datasets, \simgel{} and \simdigit{}, from 195 unique geometric primitives using the method in \autoref{sec:sim_data} for both sensor types, with background images from those datasets and our own DIGIT sensors.

% Original Table
% \begin{table}[t]
% \vspace{0.8em}
% \centering
% \caption{List of datasets used for evaluation.
% % Datasets below the line are used in \autoref{sec:b_experiments}.
% * indicates within-object split.}
% \label{tab:datasets}
% \begin{tabular}{lccc}
% \toprule
% Name & Sensor & Training \# (\# obj.) & Validation \# (\# obj.)\\
% \midrule
% \ofrgel        & \gelsight & 887 (36) & 135 (6)     \\
% \simgel        & \gelsight & 200K (135) & 113K (60)       \\ 
% \ycbdigit           & \digit    & 2578 (7) & 1084 (3)    \\
% \simdigit      & \digit    & 200K (135) & ---    \\
% % \hline
% \midrule
% \selfcollected & \digit    & --- & 547 (13)           \\
% \graspgel      & \gelsight & 13372 (6*) & 3379 (6*) \\
% \bottomrule
% \end{tabular}
% \end{table}
\begin{table}[t]
\vspace{0.8em}
\centering
\caption{List of datasets used for evaluation.
% Datasets below the line are used in \autoref{sec:b_experiments}.
* indicates within-object split.}
\label{tab:datasets}
\begin{tabular}{lccc}
\toprule
Name &  Training \# (\# obj.) & Validation \# (\# obj.)\\
\midrule
\ofrgel         & 887 (36) & 135 (6)     \\
\simgel         & 200K (135) & 113K (60)       \\ 
\ycbdigit       & 2578 (7) & 1084 (3)    \\
\simdigit       & 200K (135) & ---    \\
% \hline
\midrule
\selfcollected  & --- & 547 (13)           \\
\graspgel       & 13372 (6*) & 3379 (6*) \\
\bottomrule
\end{tabular}
\end{table}

% \subsection{Mesh-based Tactile Predictions}
% \subsection{Baseline Training and Ablations}
% basic evalution of framework
% florian: base line training and ablations
% testing out models performance between two sensor types
% sim2real problems

\subsection{Model Architecture Design Choices}\label{sec:result_design}

\begin{table}[t]
\centering
\caption{Ablation study on network design choices evaluated on the GelSight OFR dataset.
All metrics are reported on the held-out validation set.}
\label{tab:design_choices}
\begin{tabular}{l|ccc}
\toprule
Variant & PSNR ($\uparrow$) & SSIM ($\uparrow$) & LPIPS ($\downarrow$) \\
\midrule
\base{}          & 24.28 & 0.52 & 0.40 \\
\bgenc{}         & 25.52 & 0.57 & \textbf{0.34} \\
\bgsceneenc{}  & 24.91 & 0.54 & 0.38 \\
\bgstack{}      & \textbf{25.53} & \textbf{0.59} & 0.35 \\
\bottomrule
\end{tabular}
\end{table}

To validate different model design choices, we train different \flowtouch{} variants on the \ofrgel{} dataset and conduct an ablation study to assess the contribution of each individual design choice summarized in \autoref{tab:design_choices}. 
We refer to the vanilla \flowtouch{} model as \base{}. 
\bgstack{} refers to the extension of using the background stacking method as described in \autoref{subsub:back_img_as_ch_st_ctxt}.
We compare \bgstack{} to an alternative approach of using the DINOv2~\cite{oquab2023dinov} image encoder (refered to as \bgenc) and condition our \flowmatching{} process on it during training.
We also train a version that takes the DINOv2-encoded RGB image of the overall touch interaction scene as an additional input, called \bgsceneenc{}.
Quantitative results on the validation set are reported in \autoref{tab:design_choices} using three image quality metrics: PSNR, SSIM, and LPIPS~\cite{zhang2018perceptual}. 
While PSNR and SSIM are standard reconstruction quality measures~\cite{psnr_ssim}, they exhibit well-known limitations in the tactile domain, where signals are spatially sparse and sensitive to rotational and translational misalignment introduced by real-world calibration inaccuracies. LPIPS, as a learned perceptual metric, serves as a complementary measure that is more robust to such artifacts.

Baseline FlowTouch is already capable of achieving competitive performance across all metrics, which demonstrates the effectiveness of our PCN-based approach. The reported PSNR values are consistent with those by Gao~\emph{et al.}~\cite{gao_objectfolder_2023} on their visuo-tactile-cross-generation task, despite them having access to a larger training corpora and relying exclusively on simulation data.

Consistent with observations in prior work~\cite{li_connecting_2019}, we find that providing the model with an undeformed background tactile image as input yields measurable performance gains. 
All variants that make use of the background tactile (\bgstack{}, \bgenc{}, \bgsceneenc{}) outperform the unconditioned \base{} variant across all three metrics.
Notably, augmenting the model with scene-level visual context does not yield to further improvement. Additionally, \bgsceneenc{} underperforms \bgenc{} across all metrics, suggesting that the scene embedding provides a weak signal that may interfere with tactile feature learning.

Finally, \bgstack{} achieves performance on par with \bgenc{}, while avoiding the additional parameters and inference cost of the DINOv2 encoder. We also evaluated the impact of using a pretrained PCN encoder (specifically, TripoSG's~\cite{li2025triposg} encoder), but found little evidence of it outperforming the simple linear layer in our design, possibly due to the relatively small number of the PCN input. Given this parity in performance, we adopt \bgstack{} as our default configuration in all subsequent experiments.

\subsection{Training Techniques for Domain Adaptation}\label{sec:result_training}

We employed several domain adaptation techniques to bridge the sim-to-real gap in our data and to avoid catastrophic forgetting when fine-tuning on a much smaller real dataset as described in \autoref{sec:domain_adaptation}.
An ablation study of these training methods in shown in \autoref{tab:training_comparison} on the same metrics as those used for the architecture study above.
Our baseline (\naiveft{}) is trained with the pretraining/fine-tuning approach outlined in \autoref{sec:domain_adaptation}. All other methods use the baseline with only their specific technique added. The metrics are provided as differences towards the baseline and are given for three datasets. \ofrgel{} and \ycbdigit{}'s training split were used as fine-tuning datasets, while \selfcollected{} were only used for validation. All metrics are shown on their respective validation datasets.

\autoref{tab:training_comparison} shows that \jitter{} does not improve the training. Instead, it hurts all metrics compared to the baseline training. 
We hypothesize that its benefit of improving generalization does not apply here, as it obfuscates the color space which encodes important geometric data about the deformation.
The \sparshloss{} slightly hurts the \gelsight{} performances but increases the PSNR values for both \digit{} datasets.
This is expected as it optimizes in the \sparsh{} embedding space, which does not necessarily imply improving metrics computed in the RGB space.
However, as demonstrated later in \autoref{sec:result_grasp}, it can improve performance in downstream application and is therefore an important feature of our method. 

% Both \sparshloss{} and \optreset{} slighly hurt the \gelsight{} performance but improve the \digit{} performance. \tobi{@carlo some explanation why the \sparshloss{} performs as such here}. 
The \optreset{} shows slightly better values especially for the PSNR of the \digit{} datasets. In this variant, the optimizer does not experience a sudden shock when it sees new data, leading to smoother learning phase transition and increased generalization. The \domaincond{} significantly boosts performance across the board and is, thus, our best performing technique. This result was expected, as this technique helps the model to focus on the high-level similarities between the domains and lead to better generalization. % especially visible in the training curves. 

%The \domaincond{} significantly boosts performance on all metrics for all our datasets and is, thus, our best performing technique. This result was expected since it is a well established concept in domain adaptation. It helps the model to focus on the high-level similarities between the domains and, therefore, to generalize better.

During training, we noticed that doubles edges are almost perfectly learned in our simulation-only pretraining phase across all configurations, thanks to a similar geometry being present in the simulation datasets. Once the fine-tuning starts, some of these properties are forgotten and relearned after some steps, however with decreased quality. Although \domaincond{} was capable of retaining it better than others, we hypothesize these double edges have not been learned in generalized way, which is why they do not transfer well even under this condition. Additionally, injecting a similar multi-scale Gaussian noise to the ground truth PCNs as we do to the simulation-generated data improved the sim-to-real transfer.

Based on these results, we trained a final model which combines \sparshloss{}, \optreset{} and \domaincond{} which we refer to as \combined{}. Interestingly, this variant scores slightly worse than the \domaincond{} variant. However, based on the visualization of the prediction (such as \ofrgel{}, left and \selfcollected{}, right in \autoref{fig:A_viz}), we hypothesize that the combined techniques helps to retain more tactile-relevant features in the image, at the cost of the pixel-based metrics.

\begin{table*}[t]
\vspace{0.8em}
\centering
\caption{Quantitative comparison of different sim-to-real training techniques from finetuning on \ofrgel, \ycbdigit, and unseen \selfcollected{} datasets after simulation pre-training. Values for experimental methods are shown as absolute differences relative to the Naive Finetuning baseline and are calculated on the validation dataset.}
\label{tab:training_comparison}
\begin{tabular}{l ccc ccc ccc}
\toprule
\multirow{2}{*}{Training Method} & \multicolumn{3}{c}{\ofrgel{}} & \multicolumn{3}{c}{\ycbdigit{}} & \multicolumn{3}{c}{\selfcollected{} (unseen)} \\
\cmidrule(lr){2-4} \cmidrule(lr){5-7} \cmidrule(lr){8-10}
 & PSNR ($\uparrow$) & SSIM ($\uparrow$) & LPIPS ($\downarrow$) & PSNR ($\uparrow$) & SSIM ($\uparrow$) & LPIPS ($\downarrow$) & PSNR ($\uparrow$) & SSIM ($\uparrow$) & LPIPS ($\downarrow$) \\
\midrule
\naiveft{}     & $24.58$ & $0.553$ & $0.410$ & $20.23$ & $0.612$ & $0.284$ & $16.05$ & $0.355$ & $0.415$ \\
\addlinespace
% jitter
\jitter{}                  & $-0.43$ & $-0.031$ & $+0.009$ & $-0.56$ & $-0.015$ & $+0.013$ & $-0.37$ & $-0.030$ & $+0.004$ \\
\sparshloss{}              & $+0.17$ & $-0.001$ & $-0.006$ & $-0.20$ & $-0.006$ & $+0.005$ & $+0.06$ & $-0.000$ & $-0.002$ \\
\optreset{}                & $+0.28$ & $-0.005$ & $-0.004$ & $-0.03$ & $-0.001$ & $+0.000$ & $+0.07$ & $+0.001$ & $-0.001$ \\
\domaincond{}              & $\mathbf{+0.90}$ & $\mathbf{+0.029}$ & $\mathbf{-0.045}$ & $\mathbf{+2.18}$ & $\mathbf{+0.058}$ & $\mathbf{-0.055}$ & $\mathbf{+0.78}$ & $\mathbf{+0.032}$ & $\mathbf{-0.021}$ \\
\combined{}                & $+0.86$ & $+0.025$ & $-0.041$ & $+1.77$ & $+0.050$ & $-0.047$ & $+0.23$ & $+0.001$ & $-0.003$ \\
\bottomrule
\end{tabular}
\vspace{-0.5em}
\end{table*}

% \subsection{Experiments}\label{sec:b_experiments}

\subsection{Generalization Towards New Sensor Instances}
\begin{figure}[t]
  \vspace{5.1pt}
  \centering
    \includegraphics[width=1\linewidth]{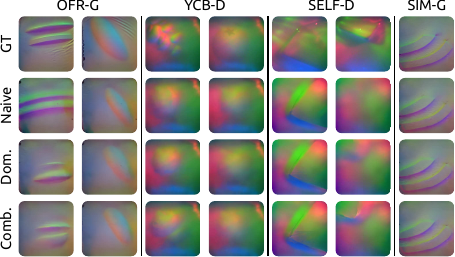}
        \caption{Tactile predictions of the model ablations listed in \autoref{tab:training_comparison}. Samples are taken from the validation dataset.}
        \label{fig:A_viz}
\end{figure}
\begin{figure}[t]
  \vspace{5.1pt}
  \centering
    \includegraphics[width=1\linewidth]{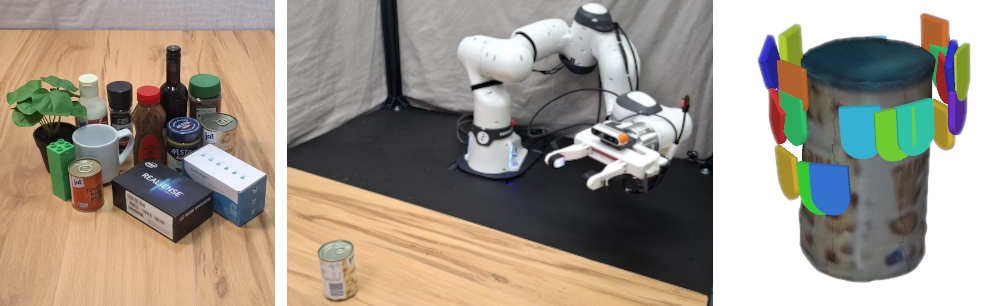}
        \caption{\textit{Left:} Objects used for data collection. \textit{Center:} Robot setup. \textit{Right:} Collected grasp poses aligned to generated mesh, with unique color for each finger pair.}
        \label{fig:eval_setup}
\end{figure}

% \subsubsection{In the Wild Experiments} % generalization towards new sensor isntances
% \subsubsection{Flowtouch - Full pipeline evaluation}
% pretraining -> zero shot
% evaluation
% this should be utn dataset digit
To test our model's ability to predict tactile images in a zero-shot manner, we collected our own dataset using the setup shown in \autoref{fig:eval_setup} on 13 different common grocery and household objects. Our setup consists of a \frthree{} robot and gripper with a 3D-printed fingers for mounting the DIGIT sensors, and a wrist-mounted D435i camera, and use Robot Control Stack~\cite{rcs} for orchestrating the setup. After fixing the object onto the table, we take a scene picture for SceneComplete by moving the robot backwards, and generate the mesh as described in \autoref{sec:meth_pointcloud}. Afterwards, we put the robot in guiding mode to collect touch samples, 537 in total. Finally, we use the robot's kinematics combined with a calibrated camera to reconstruct the touch poses around the mesh. This dataset is kept entirely for testing zero-shot generalization. For training our model, we used the \ycbdigit{} and \simdigit{} datasets described above.

% \jin{Discussion about outcome is missing}
\autoref{tab:training_comparison} shows a consistent pattern for the analyzed features on the \selfcollected{} dataset as in the previous section.
Nonetheless, compared to \ycbdigit, the metrics on our dataset are considerably worse.
Besides the obvious impact of zero-shot evaluation, we suspect that seeing a sensor instance during training allows the model to pick up on instance-specific quirks. Additionally, images from \digit{} has noticeably more artifacts than \gelsight{}, which may be another contributing factor.
Despite what the metrics might suggest, our model is still able to generalize to some extent, as seen in \autoref{fig:A_viz}.
For example, the first column of the \selfcollected{} dataset shows that the geometric shape of the touched object is roughly captured.
The second column shows an interesting observation, as it demonstrates that our \combined{} model is able to provide a sharper contact image than the \domaincond{}, while \naiveft{} fails to display anything.

\subsection{Grasp Stability}\label{sec:result_grasp}
% Not sure about the ordering. Model -> Sim -> Calculating? 
\begin{figure}[b]
  \centering
    \includegraphics[width=0.8\linewidth]{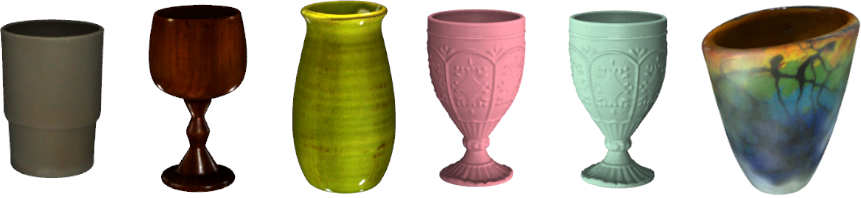}
        \caption{Meshes used for the \ofb{} grasp stability test.}
        \label{fig:ofb_obj}
\end{figure}

% Lastly we want to verify that our tactile predictions retain useful information for downstream tasks in robotic manipulation.
% We chose the task of evaluating grasp stability (TODO cite) where based on a tactile image a binary classifier has to predict if a grasp at that specific location would be successful.
% Classically (TODO cite) a robot arm would go to a specific location and close the gripper in order for the tactile sensor to produce a reading.
% A major disadvantage in this approach is the non-globality which FlowTouch completely overcomes as all tactiles can be predicted from a mesh which has to be produced only once.
% Thus our expectation is that our model does not outperform the exisiting baseline trained on the ground-truth tactiles but should rather very closely match it.
% \textbf{which model training recipe did we use? Sparsh+jitter+bg+...}
Finally, we evaluate whether our tactile predictions preserve sufficient information for downstream robotic manipulation tasks. Specifically, we adopt the grasp stability estimation task~\cite{gao_objectfolder_2023}, in which a small multi-layer perceptron-based binary classifier predicts whether a grasp would be stable enough to lift the object given the ResNet-18 encoding of a tactile image acquired at a candidate contact location. 

% \jin{this section is bit too long, could be shortened} In the conventional pipeline \cite{gao_objectfolder_2023, calandra_feeling_2025}, a robotic manipulator must physically traverse to each candidate location and close the gripper to obtain a tactile reading which requires repeated physical interaction with the object. FlowTouch fundamentally addresses this limitation: since tactile images can be synthesized for arbitrary surface locations from a single object mesh, the acquisition process needs to occur only once, enabling global coverage without additional physical probing. Accordingly, we do not expect our approach to surpass the baseline classifier trained on ground-truth tactile measurements; rather, we hypothesize that the predicted tactile images retain sufficient fidelity to yield classification performance that closely approximates the upper bound established by the ground-truth-trained model.

We use the ObjectFolder Benchmark pipeline~\cite{gao_objectfolder_2023} to collect the dataset, called \graspgel, which performs simple grasps over variations on height, rotation and $xy$-offsets in PyBullet. A tuple of RGB-D camera images and a Taxim image is collected at the moment of the grasp, and a corresponding success label is created at the end of each episode. We made minor modifications in the dataset collection script to calculate the PCN data, change the background to random ones from \ofrgel, and collected 10,000 samples for 6 objects (see \autoref{fig:ofb_obj}) from the benchmark. We balanced the dataset to consist of roughly equal number of grasping successes and failures similar to Gao~\textit{et al.}~\cite{gao_objectfolder_2021} and split the dataset in a within-objects fashion.

\begin{table}[t]
\centering
\caption{Training dataset ablations for grasp stability prediction task. Accuracies are averaged from 100 separate trainings over the test dataset (n=3379).}
\label{tab:gs_dataset}
\begin{tabular}{cccc}
\toprule
\makecell{Classifier Image\\Source} &
\makecell{Pre-training\\Datasets} &
\makecell{Fine-tuning\\Datasets} &
Acc. (\%)\\
\midrule
GT & --- & --- & 85.83 \\
\addlinespace
Comb. A& \graspgel & --- & \textbf{86.06} \\
Comb. B& Both Sims & \graspgel & 85.17 \\
Comb. C& Sims, Webs & 10\% \graspgel & 83.74 \\
\midrule
Comb. D& Both Sims & Both Webs & 81.35 \\
\makecell{No Tactile Loss} & Both Sims & Both Webs & 78.59 \\
\bottomrule
\end{tabular}
\vspace{-1em}
\end{table}

\begin{figure}[t]
  \vspace{5.1pt}
  \centering
    \includegraphics[width=0.7\linewidth]{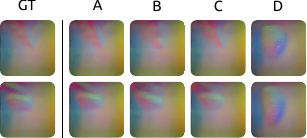}
        \caption{Tactile predictions from the grasp stability task on the dataset ablations listed in \autoref{tab:gs_dataset}. Samples are taken from the validation dataset.}
        \label{fig:C_viz}
  \vspace{-1em}
\end{figure}

%We use our FlowTouch model using \domaincond{} and \sparsh{} unless otherwise specified. % This is already done at the end of B, so not needed anymore
The dataset configurations we investigate are depicted in \autoref{tab:gs_dataset} together with the success rates of the respective experiment variants. Each configuration is trained for 6 epochs over their respective datasets.
% We introduce the dataset of tactiles which we collect in simulation and which we use later on to train the grasp stability classifier later on and denote it by \graspgel.
We first train the binary classifier on the ground truth tactile from the \graspgel{} dataset to establish our baseline variant, called \textbf{GT}, which serves as qualitative reference against our model.
All other variants are trained from the Combined variant of the \flowtouch's predictions according to different pre-training and fine-tuning configurations. 
These experiments serve to validate the output quality of \flowtouch{}, and determine the effect of different training dataset regimes. Visualizations of each variant's tactile images is shown in \autoref{fig:C_viz}.

Notably, Comb. A manages to slightly exceed the baseline's performance, while Comb. B comes very close to both of these results. We can see that Comb. C performs a bit worse, but still achieves a high accuracy at nearly 84\%.
This shows that the pretraining provides a model with strong priors that require little data to be finetuned and adapted to a new domain.
Although Comb. D achieves the worst result as expected, its 81.35\% accuracy indicates that our approach is capable of retaining important physical properties even across sensor types and domain gaps. Finally, we ablated on variant D by training the model without \sparshloss{}. This version only achieves 78.59\% accuracy, supporting our hypothesis that \sparshloss{} indeed enforces tactile information retention during training.

% Things to talk about in experiments
% 0. Training details
% 0.1. Sim dataset description - # types, total data points, etc.
% 1. Sim data validation - validating on the split set on taxim data alone
% 2. OFR data validation
%   - Regenerated the sim dataset with background images taken from the OFR dataset
%   a. sim-only, val on OFR
%   b. sim + ofr naive, val on OFR
%   c. sim + ofr reweight, val on OFR
% 3. Real world data validation
%   a. sim-only, val on real
%   b. sim + ofr naive, val on real
%   c. sim + ofr reweight, val on real
% 4. Carlo: dataset importance metric
% 5. Grasp success prediction????
%   a. sim-only, sim val?
%   b. ? datamix, validation on real
\subsection{Limitations}
Although the PCN-based tactile prediction framework shows promise, it has several limitations. First, its performance depends heavily on mesh quality. It is also highly sensitive to alignment in the robot frame, as PCN sampling requires millimeter-level accuracy that is often difficult to achieve in practice. However, ongoing advancements in mesh generation and pose estimation may mitigate these issues. 
While the model generalizes well to shapes represented in the training data, it struggles with unseen geometries. This limitation could be alleviated by scaling up the diversity and size of the simulated dataset.
Finally, our model does not explicitly encode force and instead relies implicitly on the depth of the sampled PCN to render meshes at different forces. This could be remedied by pre-processing the input PCN on a nonlinear gel deformation equation, as suggested by TACTO~\cite{Wang2022TACTO}.

\section{Conclusion and Future Work}
In this work, we introduced \flowtouch{}, a vision-to-touch framework featuring a geometry-conditioned generative prediction model. By leveraging point clouds and their corresponding normals sampled from meshes, our framework is able to effectively abstract away from scene-specific information inherent in image-only methods, and benefits from simulation data to achieve good generalization across both novel object instances and sensor types. Furthermore, we show that \flowtouch{}'s output can be used to support downstream manipulation tasks even in zero-shot settings. Finally, our framework uses scene reconstruction methods to make real-world deployment feasible.

A promising direction for future work is to incorporate textures into the model's conditioning. This will enable the prediction of high-resolution tactile features that are not captured by standard mesh generation or scanning methods, thereby making \flowtouch{} more widely applicable.

\section*{Acknowledgment}
\blackout{
This work has been partially supported by the project GeniusRobot (BMBF grant no.~01IS24083) and the Robotics Institute Germany (RIG), both funded by the German Federal Ministry of Research, Technology and Space (BMFTR).
The authors acknowledge the HPC resources provided by the Erlangen National HPC Center (NHR@FAU) under the BayernKI project no.~v106be.
This paper is supported by the DAAD programme Konrad Zuse Schools of Excellence in Artificial Intelligence, sponsored by the Federal Ministry of Research, Technology and Space.
G. K. acknowledges support by 
%the DFG under Grants DFG-SPP-2298, KU 1446/31-1 and KU 1446/32-1, as well as the
project ``Next Generation AI Computing (gAIn),'' funded by the Bavarian Ministry of Science and the Arts and the Saxon Ministry for Science and Culture.
}
%%%%%%%%%%%%%%%%%%%%%%%%%%%%%%%%%%%%%%%%%%%%%%%%%%%%%%%%%%%%%%%%%%%%%%%%%%%%%%%%

\bibliographystyle{IEEEtran}
\bibliography{bibliography.bib}

\end{document}